\documentclass[a4paper,conference]{IEEEtran}
\ifCLASSINFOpdf
\else
\fi
\DeclareUnicodeCharacter{2061}{}
\usepackage{graphicx}
\usepackage{cite}
\usepackage{amsmath,amssymb,amsfonts}
\usepackage{algorithmic}
\usepackage{textcomp}
\usepackage{xcolor}
\usepackage{graphicx}
\usepackage{caption}
\usepackage{float}
\usepackage{subcaption}
\usepackage{booktabs}
\usepackage{multirow}
\newcommand\blfootnote[1]{%
  \begingroup
  \renewcommand\thefootnote{}\footnote{#1}%
  \addtocounter{footnote}{-1}%
  \endgroup
}

\begin{document}
%
\title{TeLCoS: OnDevice Text Localization with Clustering of Script}

\author{
\IEEEauthorblockN{Rachit S Munjal*}
\IEEEauthorblockA{OnDevice AI\\
Samsung R\& D Institute\\
Bangalore, India\\
rachit.m@samsung.com}
\and
\IEEEauthorblockN{Manoj Goyal*}
\IEEEauthorblockA{OnDevice AI\\
Samsung R\& D Institute\\
Bangalore, India\\
manoj.goyal@samsung.com}
\and
\IEEEauthorblockN{Rutika Moharir}
\IEEEauthorblockA{OnDevice AI\\
Samsung R\& D Institute\\
Bangalore, India\\
r.moharir@samsung.com}
\and
\IEEEauthorblockN{Sukumar Moharana}
\IEEEauthorblockA{OnDevice AI\\
Samsung R\& D Institute\\
Bangalore, India\\
msukumar@samsung.com}
}

\maketitle

\begin{abstract}
Recent research in the field of text localization in a resource constrained environment has made extensive use of deep neural networks. Scene text localization and recognition on low-memory mobile devices have a wide range of applications including content extraction, image categorization and keyword based image search. For text recognition of multi-lingual localized text, the OCR systems require prior knowledge of the script of each text instance. This leads to word script identification being an essential step for text recognition. Most existing methods treat text localization, script identification and text recognition as three separate tasks. This makes script identification an overhead in the recognition pipeline.

To reduce this overhead, we propose TeLCoS: OnDevice Text Localization with Clustering of Script, a multi-task dual branch lightweight CNN network that performs real-time on device Text Localization and High-level Script Clustering simultaneously. The network drastically reduces the number of calls to a separate script identification module, by grouping and identifying some majorly used scripts through a single feed-forward pass over the localization network. We also introduce a novel structural similarity based channel pruning mechanism to build an efficient network with only 1.15M parameters. Experiments on benchmark datasets suggest that our method achieves state-of-the-art performance, with execution latency of 60 ms for the entire pipeline on the Exynos 990 chipset device.
\blfootnote{* Primary authors}
\end{abstract}


%
\IEEEpeerreviewmaketitle

\section{Introduction}
There has been a significant rise in multimedia content generation in recent times, due to the images captured by unconstrained cameras and scene images taken daily with added filters and effects. To analyze the diverse variety of image content, it is necessary to understand the semantic information embedded within such images. The problem of text detection in scene images and its extraction from unstructured data has gained much attention in the computer vision field owing to its wide variety of applications such as receipt scanning, visual question answering, keyword based image hashing and many others. It has become a difficult task, due to the complex scenarios such as textured background, curved orientations of text and multiple scripts present in a single image
(hitherto referred to as multi-script images) such as shown in Fig 1.
\begin{figure}[htbp]
\includegraphics[width=90mm]{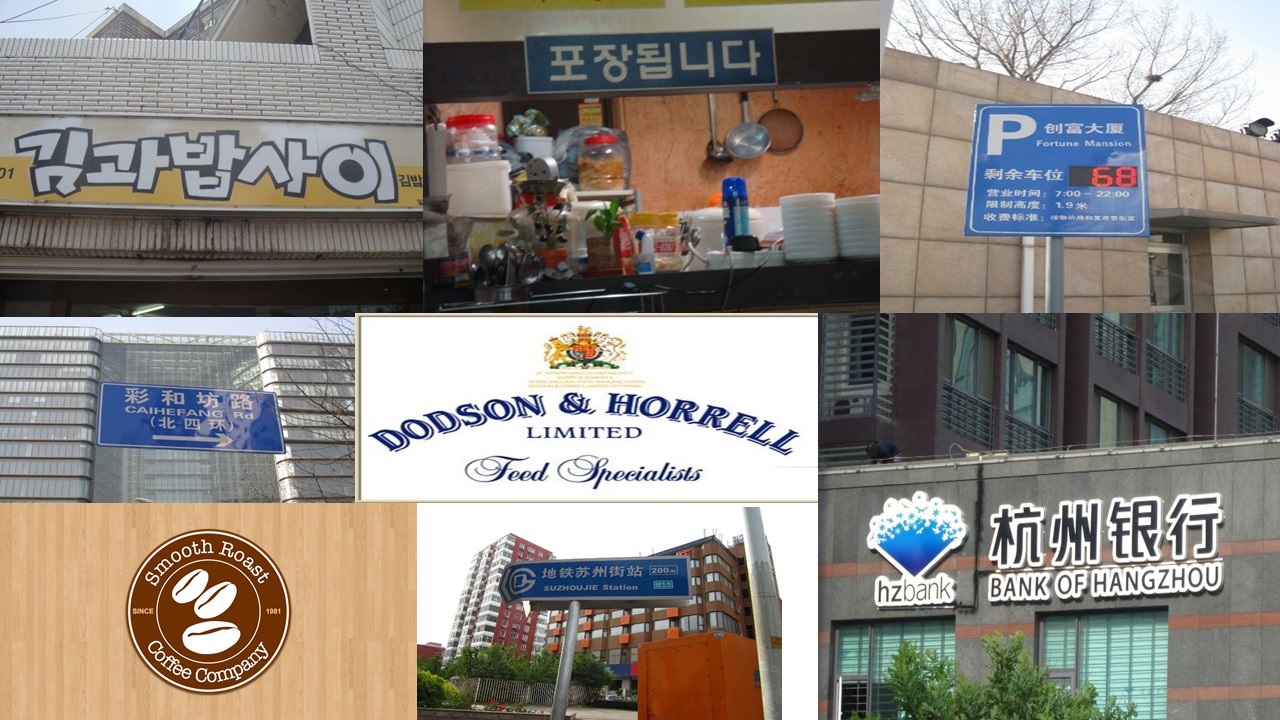}
\label{fig:label}
\centering
\caption{Examples of multi-oriented, multi-script scene text}
\centering
\end{figure}

Existing OCR systems assume the scripts for the text to be known in advance and do not efficiently handle mixed-script images. Some solutions perform script identification at the line-level post which the lines are passed to the script specific recognition model. This approach however poses two important limitations. First, they perform script identification at a line level, i.e. they assume a text line to be in a single script. Second, the entire OCR process is broken down into three major subcomponents: text localization, script identification and script specific text recognition. Each of which are developed independently, but used in sequence.

\begin{figure*}[htbp]
\includegraphics[width=190mm]{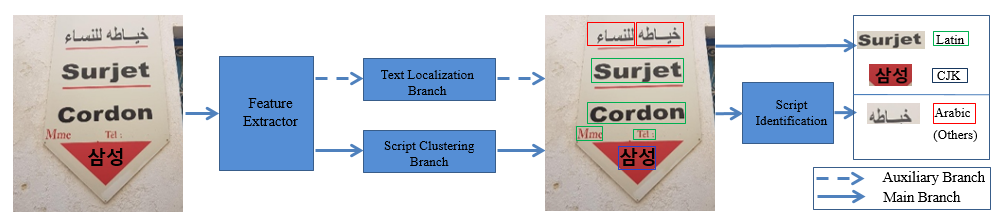}
\label{fig:label}
\centering
\caption{End-to-End TeLCoS Pipeline Diagram: Dual branch multi-task learning architecture which performs Simultaneous Text Localization and High-Level Script Clustering into major script groups. Words from 'Other' class are fed to the Script Identification module for further script detection.}
\centering
\end{figure*}

In text detection, usually a convolutional neural network is used to extract feature maps from a scene image and then different decoders are used to decode the text
regions. While, in script identification, a network for sequential prediction is conducted on top of text regions one by one. For multi-script document recognition several calls to the script identification module are required. This incurs a massive computational overhead, making its usage futile in the case
of real world solutions. Also, separate learning of the two tasks fails to capture the correlation between text detection
and script identification.

A multi-lingual text recognition system should be able to handle arbitrary mixed script content.  Generally, a significant amount of multi-lingual text images contain the most widely used scripts such as Latin or Chinese-Japanese-Korean (CJK) in the majority with some other scripts. Also, most images are bi-lingual or tri-lingual in nature. Taking advantage of these facts, we propose TeLCoS: OnDevice Text Localization with Clustering of Script, a multi-task convolution neural network that performs simultaneous text localization and an initial high-level script clustering into Latin, CJK and Others. The leftover word regions and groups whose script is unknown can then be passed for individual script identification as per the requirement. This gives a huge computational gain compared to carrying out independent script identification for each text word present in the image. The TeLCoS network is end-to-end trainable and learns features that are shared between these two tasks. The shared features reduce the feature extraction to a single network as shown in Fig 2.

For training, we used ICDAR 2019 [6], Multilingual Scene Text (MLT 2017) [20] and the MLe2e [26] dataset. Additionally, to manage the inadequate number of training samples for certain scripts, we introduce a fast and efficient contour based method for generating synthetic images which resemble semantics of natural images to improve the performance on the varied input scenarios.  

 Our model has 1.15M parameters, which is at least 50x less than other top models in this field with comparable accuracy. It manages to achieve state of the art performance, in both the tasks of text detection and script clustering and takes under 60 ms for the entire pipeline on Exynos 990 chipset device. We achieve an H-Mean of 90.5\% for text detection on the IC13 dataset.

The main contributions of this paper are summarized as follows: 
\begin{itemize}
\item We propose an end-to-end trainable framework for simultaneous text detection and script clustering which leads to a decrease in the computational overhead of a separate script identification module.
\item We propose a novel Structural similarity based channel pruning method as a compression technique to generate models suitable for inference on resource constrained devices.
\item We introduce an efficient method for generating synthetic images which resemble semantics of natural images to improve performance on all input scenarios. 
\end{itemize}

\section{Related Work}

\subsection{\bf Text Localization}
Most solutions of text localization fall under two main categories, one that is based on identifying the connected components in the system, with subsequent tagging of each of the components as text or a non-text region, and the other that is based on sliding a variable scaled window depending on input dimensions throughout the image, to identify in each iteration whether the part of the image under the current purview of the window is a text region or not.  

Sliding window based classifier can be found in the works of Wang et al. [2] where at each instance of a sliding image view, traditional image features, like HOG descriptors are computed, and then those features computed, are fed into a modified Random Ferns Classifier [3], to estimate whether or not that window has characters in it. An extremely important feature, that was often extracted and fed into most text based classifiers, was the Stroke width classifier introduced by Epshtein et al [4].

The emergence of Stroke width transform led to a surge of algorithms that took advantage of the said feature, and one specific idea that combined Maximally Stable Extremal Regions (MSER) and geometry based features of Stroke width, to feed into classifiers as input, and get the text regions in the image presented by Yin et al [5] gained a lot of traction.

With the advent of deep learning methods, a new breed of solutions started gaining popularity, which again was classified into two categories based on the way output was presented in each of these methods viz. Semantic Segmentation based  and bounding box regression based. Generally, both of these methods, focussed on finding words, and not just individual characters in images. 
One of the most popular bounding box based Neural Networks solution was published by Liao et al, called TextBoxes, which has over the years been improved iteratively (Textboxes++ [8]). The solution in this paper, for the first time, predicted the text regions in an image, post a single pass over the network. Other such networks were proposed as well, but all of these had a major limitation, since limiting output to be in the form of boxes excludes a lot of multi oriented text that could occur in texts.  

Thus, there was more focus on semantic segmentation as an output mechanism, and one such algorithm proposed by Dafang et al [9], helped overcome this limitation and provided the feature map which gives bounding box coordinates after some post processing.  The current SOTA model of CRAFT [1] that is also the inspiration for the architecture adopted for our joint task, also produces output in similar way. But this architecture has a very high number of parameters, which makes it unsuitable for on-device inferencing.

\subsection {\bf Script Identification}

Till now, Script identification has been treated as a separate problem altogether. Two different types of approaches have been widely used. Earlier, statistical approaches have been used to obtain texture level features from Local Binary Patterns [10] or rotation invariant multi-channel Gabor filters [11].

Above mentioned methods detect edges of text regions, which cannot be done for scene text images due to their complex backgrounds. So, approaches were developed with deep learning methods which took advantage of effective feature extraction using CNN. Multi-stage Spatially-sensitive Pooling Network (MSPN) method was introduced by Shi et al.[12] where they provided the first real scene text images dataset for script identification.

\subsection {\bf Joint Text Localization and Script Identification}

As above mentioned, text localization and script identification have been considered separate problem statements till now. In the above methods, every word in the original image needs to be fed into these network individually. Thus, an image containing 300 words would require more than 300 calls to script identification module during inference time. The overall computational cost required to finally complete both script identification and text localization becomes intractable in real time applications. 

A combination of a few methods listed above would have been a feasible solution, if the time restrictions are lifted. However, if any real time application that handles an unrestricted domain of images is desired, a design that performs inference in a solitary feed forward pass over the network, is a necessity, and as of today to the best of our knowledge, no publicly known record of an attempt to do so exists. In our approach, we perform the high-level script clustering in conjunction with the localization of text. This helps us to identify the text location of some of the majorly used script groups initially, which can then be passed to a specialized script identification model to identify the finer differences within scripts of a particular group as per the requirement. As this second pass to script identification can be skipped for most of the localized words, it helps us reduce the total execution time considerably.

\section{Dataset Creation}

For training of the end-to-end pipeline we use ICDAR 2019 [6], ICDAR 2017 [20] and MLe2e[26] datasets. Detailed description of the above datasets is given in section VII. Additionally, to enhance the diversity of the training sets we create 4L synthetic images. We use the Open Image dataset [13] as background images. For synthetic text generation, we propose a fast and efficient contour based approach to render text on natural images. Initially, we filter out images with indoor and outdoor scenarios and remove the images with any kind of text already present in the image. This is done to avoid true negatives in the ground truth, due to the lack of proper character/word level annotations for text present in such images. Then, to render text on the filtered background images, we apply a channel-wise median blur to remove the small edges followed by canny, to extract the edge and contour information. The blur kernel size is kept tunable depending on the font size and width. The image is then split into multiple patches of variable size, on the basis of the contour per unit area information. Finally, we place the text at word, line and para level, in regions of low contour density. We use text blending with the background and other transformation techniques like perspective, sine or affine transform, rotation, sharpening or smoothening. The sample synthetic images generated by this method are shown in Fig. 3. This method doesn't require any additional information related to the image like segmentation map, depth map etc. which were required for SynthText [7].
 
\begin{figure}[t!]
\includegraphics[height=55mm]{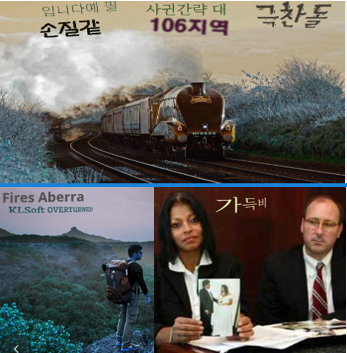}
\label{fig:label}
\centering
\caption{Synthetic Images Generated using contour based approach }
\centering
\end{figure}

For text annotation, we generate a pixel-level character and affinity score using the method followed by CRAFT [1].  This is done by encoding the probabilities of character centers and space centers between adjacent characters respectively. Post this every pixel is given a script score, on the basis of the script group present in the region, if either of the affinity or region score for that pixel is greater than a set threshold. All the score maps are modelled as probabilistic Gaussian distributions.

\begin{figure*}[t!]
\includegraphics[width=180mm]{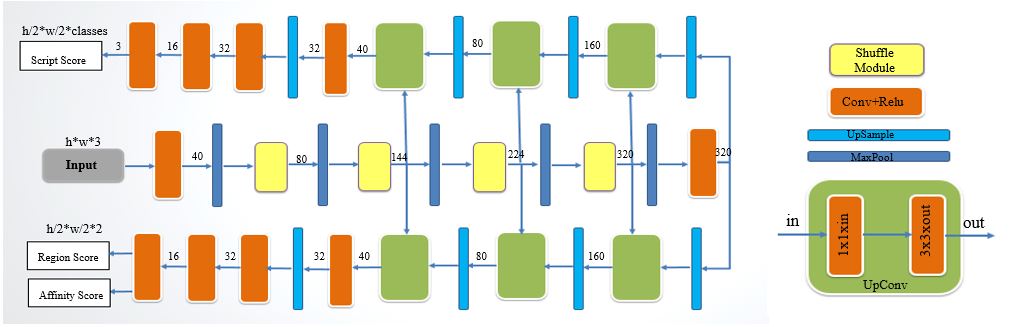}
\label{fig:label}
\centering
\caption{TeLCoS Network Architecture Overview: Skip Connections from feature extractor with ShuffleNet Blocks are fed to both the text localization and script clustering branch. Script Clustering Branch outputs pixel-wise script scores for different script groups. The localization branch is an auxillary branch which is used only at training time to improve word boundaries by learning character spaces, using affinity and character region scores.   }
\centering
\end{figure*} 

\section{Our Approach}

TeLCoS is an end-to-end trainable framework which shares convolution features between the two tasks of text detection and script identification. It supports the detection of multi-oriented, multi-scale scene text and incidental text, as well as parallely performs a high-level script clustering. We model the Text Localization and Script Clustering task as a semantic segmentation task, treating words of different script groups as distinct object classes. The clustering is done into the majorly
used script groups, i.e Latin, CJK and Others. The leftover word regions and groups whose script is classified as ’Others’
is then passed for individual script identification as per the
requirement.

We propose a light weight CNN based multi-task dual branch network with a shared feature extraction backbone. For feature extraction, we use a series of shuffle modules, due to its low memory footprint and fast inference speed. It is followed by a convolution layer which is fed to a dual branch U-Net architecture to obtain multiscale feature information, to strengthen the network’s feature extraction ability. The overall architecture for our framework can be seen in Fig. 4.

\subsection{Feature Extractor}
We wish to establish an effective mechanism for sharing relevant information across the two tasks of text detection and script clustering. In addition, we want our end-to-end pipeline to have low memory footprint and low latency. In our architecture, we share features of the initial layers after which the sharing is blocked, resulting in task-specific sub-networks for branches beyond it. The results of script clustering accuracy when trained alone for script identification task vs when trained jointly along with the text localization task are shown in TABLE VI. Hence, we are able to achieve model compression and acceleration by considering task relatedness.

For the feature extraction backbone, we use the Shufflenet V2 [14] blocks. A basic Shuffle block splits the incoming channels into two parts, in which one split is passed as it is, to retain the low-level features. Whereas, depthwise convolution is applied on the second split. The splits are then concatenated and shuffled. We experiment with other light weight feature extraction components for the backbone, like Inverted Residual blocks of MobileNetV2 [15] and fire modules of SqueezeNet [16]. We observe that shuffle modules perform better in terms of speed and accuracy, due to its low-level feature retention property and channel shuffling. We achieve latency of 60 ms on Exynos 990 as shown in Table I.

\begin{table}[b!]
\centering
\caption{ Text Localization Accuracy and Inference Speed Comparison with different Feature Extractors on IC13 Dataset}
  \label{TLfeaturecomparetable}
  \begin{tabular}{cccccc}
    \toprule
    \multirow{2}{*}{BackBone} &   
     \multirow{2}{*}{Precision} &
     \multirow{2}{*}{Recall} &
    \multirow{2}{*}{H-Mean} & 
    \multirow{2}{*}{  Speed(ms)} &  \\ & \\
 \midrule
  \multirow{1}{*} {MobileNetV2 Blocks} & 91.62 & 81.82 & 86.4 & 78\\
  \multirow{1}{*} {SqueezeNet Fire Modules}  & 90.25 & 80.78 & 85.3 & 73\\   
    \multirow{1}{*} {ShuffleNetV2 Blocks}& \textbf{93.56} & \textbf{86.72} & \textbf{90.5} & \textbf{60}\\
\bottomrule
  \end{tabular}
\end{table}

\subsection{U-Net}
The first branch is a regression branch responsible for learning the character and affinity distributions of the text. The affinity scores from the regression branch help to gain an understanding of word boundaries accurately and regulate the model from joining nearby words. The other is a classification branch which learns the class group of each text region. The high-level grouping is done by splitting the scripts with similar looking structure and features together. The division of groups is as follows: Latin, CJK (Chinese, Japanese and Korean Scripts) and Others (Arabic, Indic and other common scripts). This grouping helps to reduce the bottleneck time of applying script identification on each word crop individually. For instance, the word regions identified as Latin can be directly passed to the recognition model for the Latin language. Similarly, a combined recognition model for the CJK language is also common, which circumvents the need for further bifurcation of the CJK group. The word region classified as ‘Others’ is passed to the script identification model for further script detection.

Each branch of the U-Net consists of 3 UpConv and UpSample blocks, in which skip connections are fed from the 2nd, 3rd and 4th shuffle module, to support multi-scale text detection. Each UpConv block consists of a 1x1 convolution, followed by a 3x3 convolution. The UpConv blocks are followed by a resize bilinear layer for upsampling, with upscaling factor of 2.  The output node of the text localization branch is of size h/2 * w/2 * 2, where the 2 channels predict the pixel-level character and affinity scores. Whereas, the output of the clustering branch is of size h/2 * w/2 * 4, the channels denoting the four classes i.e. Latin, CJK, Other, None.

We train the regression branch as an auxiliary branch, to assist the script branch in learning word boundary information and we remove it from the graph while inferencing, to reduce computational cost. Though the script branch alone is capable to perform both the tasks, i.e. Text localization and Script Clustering, empirically it is observed that when trained in conjunction with the regression branch, both the tasks perform better because of the shared features and joint optimizations, indicating that the knowledge from different tasks complements the joint training. Due to the lack of affinity score information, some adjacent words start joining, which leads to a drop in accuracy when trained without the auxiliary branch.

\subsection{Knowledge Distillation}
Real-time text detection on mobile devices require very light-weight models and minimal parameter networks, while maintaining a base level of accuracy. We apply knowledge distillation to train a smaller student network, with the help of a fully trained similarly structured teacher network. We try to transfer knowledge learned by a complex and accurate teacher network to a small and compact student network. The soft targets of the teacher network i.e. predicted probabilities for all classes, contains a lot more salient information hidden than just the ground truth hard targets. So, rather than searching in the entire target space, the student network is made to search in a teacher-restricted target space. 

We use the teacher network, to train a
compact student network and fine-tune it on ground truth hard targets. With knowledge distillation we get a 55\% compressed network without any noticeable loss in accuracy. Just training the student network alone with ground truth, we observed around 3.2\% drop in accuracy, whereas when trained with soft probabilities from the teacher network and fine-tuned on ground truth, the drop in accuracy was less than 0.5\%. The accuracy and size comparison for the networks is given in Table II.

\subsection{Structural Similarity based Channel Pruning}
Deep neural network models tend to produce a lot of redundant features and suffer from heavy over-parameterization, so there is always a scope of pruning the network to reduce the model parameters and inference time, while maintaining the same level of accuracy. In recent works, different pruning techniques have been proposed. For weight pruning the weights are ranked according to their magnitude and k percent of them are set to zero, while for neuron pruning entire columns of the weight matrix are set to zero according to their L2-norm.   

Many weights in a network have very small value and hence weight pruning can be done effectively to reduce the model size. But, due to poor cache locality caused by random connectivity, the speed acceleration is
very limited. Also, currently there is no support for optimizations to reduce inference time for the sparser networks in many deep learning libraries. 

As our main focus is to obtain faster inference speed on-device, we shift our focus to channel pruning. We propose a novel structural similarity based pruning and tackle the task of channel pruning in a slightly different way than [17], [18] which compares the filters of each layer. Instead of the filters of the current layer, we focus on the input being passed to the next layer. We prepare a small set of the representative dataset, which covers a wide range of possible input images. Then, the images are fed to the network and the output of activations are fetched from the layers to be pruned. For each channel-wise output of activation, we calculate the structural similarity index with all other activations of the same layer, on all the images from the representative dataset.

We group together the channels with high SSIM score. The SSIM threshold is set to around 0.4 to 0.7 depending on the layer to be pruned and percentage of pruning required. From each set around k percent of channels are removed. 

\begin{table}[b!]
\centering
\caption{Comparison of parameters and accuracy after feature pruning on IC13 dataset}
  \label{TLpruningtable}
  \begin{tabular}{cccccc}
    \toprule
    \multirow{2}{*}{Layer} &   
     \multirow{2}{*}{Teacher} &
     \multirow{2}{*}{Student} &
    \multirow{2}{*}{Channel Pruning} &  \\ 
    \multirow{2}{*}{} &   
     \multirow{2}{*}{\bf (TeLCoS-Plus)} &
     \multirow{2}{*}{\bf (TeLCoS-Stud)} &
    \multirow{2}{*}{\bf (TeLCoS)} &  \\ &
     \\
 \midrule
\multirow{1}{*} conv1 (5x5) & 64	& 48 &	40 \\
\multirow{1}{*}shuffle1	&64, 128	&48, 96	&40, 80\\
\multirow{1}{*}shuffle2	&128, 224	&96, 172&	80, 144\\
\multirow{1}{*}shuffle3	&224, 384	&172, 264&	144, 224\\
\multirow{1}{*}shuffle4	&384, 512	&264, 344&	224, 320\\
\multirow{1}{*}conv2 (3x3)	&512	&344	&320\\
\multirow{1}{*}upconv1	&512, 256	&344, 172&	320, 160\\
\multirow{1}{*}upconv2	&256, 128	&172, 96	&160, 80\\
\multirow{1}{*}upconv3	&128, 64	&96, 48	&80, 40\\
\multirow{1}{*}conv3 (3x3)	&48	&40	&32\\
\multirow{1}{*}conv4 (3x3)	&32	&32	&32\\
\multirow{1}{*}conv5 (1x1)&	16&	16	&16\\
\multirow{1}{*}conv6 (1x1)&	2 & 2 &	2\\
\midrule
\multirow{1}{*} {\bf Total Params} &	3.1M & 1.4M & 1.15M\\	
\multirow{1}{*} {\bf Accuracy} &	90.7\% & 90.3\% & 90.5\%\\	
\bottomrule
  \end{tabular}
\end{table}

\begin{figure}[htbp]
\includegraphics[width=70mm,height=70mm]{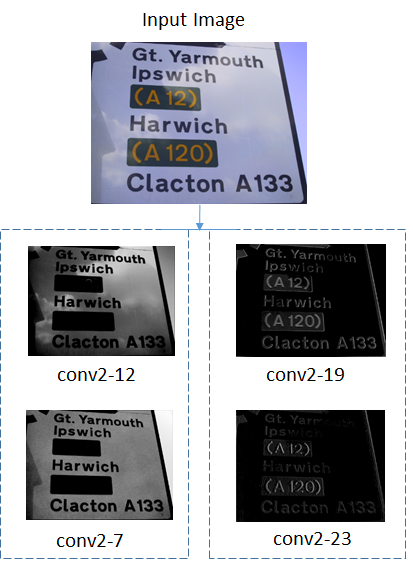}
\label{fig:label}
\centering
\caption{Channel-wise sample activations from the 2nd convolution layer. Activations with similar SSIM scores are grouped together, and x\% channels are dropped from each group.}
\centering
\end{figure}

For finding removal candidates from each set, some approaches are 1) Weight Sum: Remove filters with lowest absolute sum of kernel weights and 2) Average percentage of
zeros (APoZ): Removal based on sparsity of activations. The weight sum method is known to have a poor accuracy, since it only takes the magnitude of kernel weights into consideration. Also, in an already lightweight network like ours, the features with high APoZ score is usually low.

We utilize the pruning based on APoZ, to remove x percent channels from our structurally similar subsets. This is followed by removing the top (k-x) percent channels based on highest SSIM score within each subset. This helps in removal of a subset of filters which are capable of producing similar outputs on a wide set of data. The idea behind this is, that filters producing similar activations can be grouped together and a few redundant features from each group can be dropped. The contribution made by these features can be quickly retrieved by other features of the group.

The removal of channels from each layer, might need some re-adjustment in weights. So, the selected channels are pruned from the current layer and the input of the next layer, and the rest of the network is re-initialized from the saved weights and fine-tuned. Pruning all layers at once impairs the network and makes it difficult to recover. So, we follow an iterative process to prune 2-3 layers at a time, recover network by fine-tuning, prune again. 

After the pruning process, we are able to reduce the network by 20\% from the student network, with a slight increase in
validation accuracy due to better generalizing capability of the pruned model. Table II shows the parameter reduction after performing knowledge distillation and feature pruning. The pruned network gives an accuracy of 90.5\% on IC13 dataset with only 1.15M parameters.

\subsection{Script Identification Module}
The word boxes which require further script identification are fed to the script identification module. It takes the mapped features extracted by the 2nd convolution of the text localization model as input and aims to find the script for the given text instance. We draw inspiration from MSPN architecture, using Spatially Sensitive Pooling layers after the convolution
layers and further concatenating response maps at different abstraction layers such that they describe the text at both high and low level. We apply up-down convolutions in the network, so that only the most
important features are passed to the following layers. At the end of the convolution stack, fully-connected layers need to be used for script classification. 
Due to the high memory requirement of these dense layers, we replace it with projection layers [19] in the network. This uses Locality Sensitive Hashing (LSH) based projection method. As illustrated in Fig. 6, we use a distillation training framework, which minimizes a combination of Lt and Lp which helps the projection
network learn from the larger Teacher Network. The yellow colored blocks indicate
the projection layers.

\begin{figure}[b!]
\includegraphics[width=90mm]{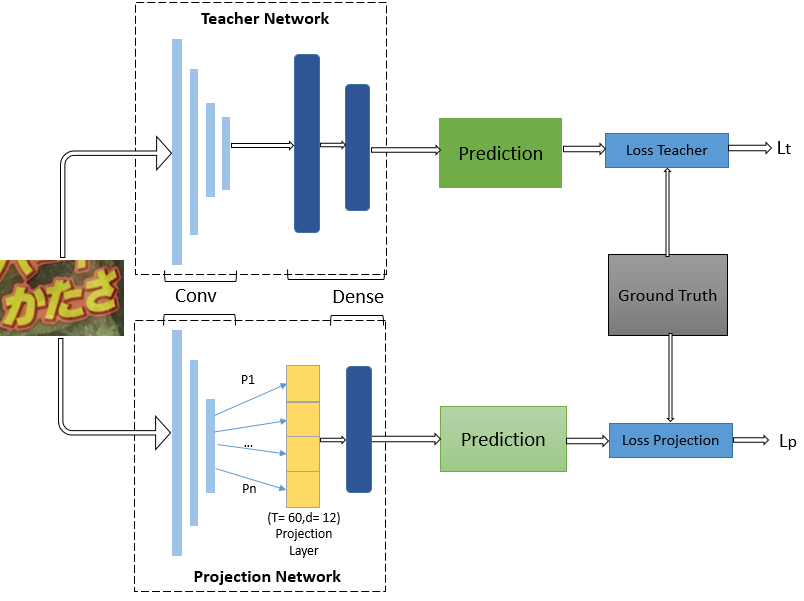}
\label{fig:label}
\centering
\caption{Script Identification with distillation training framework. The projection functions P1..Pn transforms the output of the last convolution before feeding it into the dense layers.}
\centering
\end{figure}

\begin{table*}[t!]
\centering
\caption{Comparison with other results on ICDAR 17, ICDAR 13 and MSRA-TD500 datasets for text detection task with percentage scores. P, R, H correspond to Precision, Recall and H-Mean respectively.}

  \label{our-result-table}
  
  \begin{tabular}{ccccccccccccccccccccccc}
    \toprule
    \multirow{1}{*}{Method} & &
    \multirow{1}{*}{ICDAR 17}&&&&
    \multirow{1}{*}{ICDAR 13}&&&&
    \multirow{1}{*}{MSRA-TD500}&&&
   \multirow{1}{*}{FPS}\\
    \multirow{2}{*} {}&
    \multirow{2}{*}{P} &
    \multirow{2}{*}{R} &
    \multirow{2}{*}{H} & 
    \multirow{2}{*} {}&
    \multirow{2}{*}{P} &
    \multirow{2}{*}{R} &
    \multirow{2}{*}{H} & 
    \multirow{2}{*} {}&
    \multirow{2}{*}{P} &
    \multirow{2}{*}{R} &
    \multirow{2}{*}{H} & \\ \\
 \midrule
    \multirow{1}{*} {CRAFT}  & 80.6 & 68.2 & 73.9 && 93.1 & 97.4 & 95.2 && 88.2 & 78.2 & 82.9 && 8.6\\
    \multirow{1}{*} {Corner Localization}  & 74.3 & 70.6 & 72.4 && 92.0 & 84.0 & 88.0 && 87.6 & 76.2 & 81.5&& 5.7 \\
     \multirow{1}{*} {PAN}  & 80 & 69.8 & 74.3 && - & - & - && 84.5 & 83.8& 84.1 && 26.1\\
    \multirow{1}{*} {FOTS}  & 80.95 & 57.51 & 67.25 && - & - & 87.3&& - & - & -&& 23.9 \\
	\multirow{1}{*} {SegLink}  & - & - & - && 87.7 & 83 & 85.3&& 86 & 70 & 77 && 8.9 \\
 \midrule  
 \multirow{1}{*} {TeLCoS} & 78.7 & 64.9 & 71.12 && 93.56 & 86.72 & 90.5 && 87.2& 76.9 & 81.72 && 83.9 \\
\bottomrule
  \end{tabular}

\end{table*}

\section{Optimization Function}
Applying an optimization function directly on pixel-wise deviation in predicted and actual scores may not be suitable because the character and affinity scores at every pixel are highly inter-related with its neighbouring pixels. To tackle this dependence in nearby pixels, we define a novel: logcosh-Pool (Ltl) cost function given by:

\[L_{tl}(y,y^p)=\sum_{(i=1)}^Nlog⁡(cosh⁡(y_{avgpool(s=2)}^p-y_{avgpool(s=2)}))\]

The pooling allows to avoid and cancel out the possible accumulation of small pixel level deviations in neighbouring scores which are irrelevant to us, as the smallest objective is to predict correct character boundaries instead of predicting exact pixel level scores. The function also gives a smoothening effect to the final output and a boost in the accuracy, giving lesser cuts in word boundaries. Categorical class entropy loss (Lsd) on pixel wise softmax is used for script class detection and is given by:

\[L_{sd}(y,y^p )=-\sum_{(j=0)}^M\sum_{(i=0)}^N(y_{ij}*log⁡(p_{ij})) \]

\section{Training and Implementation Details}

We carry out the training process in two steps. First synthetic dataset is used for 5 epochs and then the benchmark datasets are used for finetuning the model. Data augmentation techniques are used to ensure diversity of dataset to make it robust to any input image. First the images are resized to 768x768 pixels while maintaining the aspect ratio. Then various distortions such as rotation, gaussian blur, image saturation, graying and variation in brightness are added. The network is implemented in tensorflow 2.3 and trained on Nvidia GeForce GTX 1080 Ti with 16GB memory. We use Adam optimizer with momentum term 0.01. The initial learning rate is 0.01 and it is halved after every epoch, with early stopping mechanism.

\section{Experiments And Results}
We evaluate our suggested approach for both the defined tasks i.e. Text Localization and Script Clustering in two steps. First, the text detection results are measured using the Intersection-Over-union (IoU) metric. The detected bounding box is considered true positive if it has an IoU greater than 0.8 with the bounding box in the ground truth. Second, for script clustering, the predicted script is verified against the ground truth.
\subsection{Text Localization Evaluation}
To evaluate Text Localization accuracy, we use ICDAR2017 [20], ICDAR2013 [21] and MSRA-TD500 [22] datasets.

{\bf ICDAR2017} known as IC17 contains 900 testing images. It contains multilingual scene text data in 9 different languages.

{\bf ICDAR2013} known as IC13 contains 233 testing images. It contains focussed scene text data in English Language. Word level annotations are provided in reactangular shape.

{\bf MSRA-TD500(TD500)} contains 200 testing images. The images contain data of two languages i.e. English and Chinese.

As mentioned in above approach, we take the model output from script clustering branch as text localization is auxialary branch. After post processing of model output, we retrieve boxes corrdinates of text localization. While inferencing, images are resized to 768 max dimension maintaing aspect ratio. Even though we don't run on higher dimensions, but we achieve comparable results with other state of the art methods as shown in Table III. As our model is very light and capable for resource constrained devices, we achieve very high FPS as compared to other state of the art methods. 

\subsection{Script Clustering Evaluation}
We evaluate predicted script accuracy from our proposed approach with CVSI-15 [27], MLe2e [26] and ICDAR2017 [20] datasets.

{\bf ICDAR2017:} IC17 dataset contains 16255 words for validation set for the script clustering task which contains 9 different languages. As  English, French, German, Italian lanuages share Latin script, we assign all these to Latin class. In the same way, Chinese, Japanese, Korean are assigned to the CJK script class. While, the remaining languages are categorised as Other class. Table IV shows script clustering accuracy on IC17 dataset.
\subsection{Script Identification Evaluation}

{\bf MLe2e} is a Multi-Language dataset which contains 711 images constituting 1178 word instances in of four different scripts, namely Latin, Chinese, Kannada and Hangul. The comparison result of script detection after end-to-end pipeline is shown in Table V.

\begin{table}[b!]
\centering
\caption{Accuracy of script clustering module on IC17}
  \label{our-result-table}
  \begin{tabular}{cccc}
    \toprule
    \multirow{2}{*}{Dataset} &   
     \multirow{2}{*}{Accuracy} &
    \multirow{2}{*}{} & \\ \\
 \midrule
    \multirow{1}{*} {Latin}  & 94.9 \\
     \multirow{1}{*} {CJK}  & 91.4\\
 \multirow{1}{*} {Other}  &  92.3\\
\bottomrule
  \end{tabular}
\end{table}

\begin{table}[b!]
\centering
\caption{Comparison with other SOTA methods for end-to-end script detection on benchmark datasets}

  \label{our-scriptresult-table}
 
  \begin{tabular}{cccccccccc}
   \toprule
    \multirow{2}{*}{Method} &   
   \multirow{2}{*}{CVSI-15} &
    \multirow{2}{*}{MLe2e} &
    \multirow{2}{*}{ICDAR 2017} & \\ \\
 \midrule
   \multirow{1}{*} {Gomez [23]}  & 97.20 & 94.4 	& 86.46 \\
	\multirow{1}{*} {Bhunia [24]}  & 97.75 & 96.70 	& 90.23 \\
	\multirow{1}{*} {Patch Aggregator [25]}  & 86.0 & - 	    & 89.42 \\
	\multirow{1}{*} {Gomez et al. [26]}  & 95.11 & 91.12 	& - \\
	\multirow{1}{*} {CVC-2 [27]}  & 96.0 & 88.16 	& - \\
\midrule          
	 \multirow{1}{*} {TeLCoS} & 96.42 & 92.71 & 87.29  \\
         
\bottomrule
  \end{tabular}
 \end{table}

As illustrated in Table VI, we are able to solve a real world problem of resource constrained devices in which script detection becomes an overhead for text recognition pipeline. We are achieving script information in text localization time frame without any major drop in accuracy of identified scripts. Fig. 7 shows example of the results.

\begin{table}[t!]
\centering
\caption{Comparison of Inference time, Model size and Script wise accuracy of TeLcoS and a two staged method consisting of independent text localization and script identification module (Two stage TL+SD) on a sample image consisting of approximately 100 words.}
\label{tlsdjointcomparetable}
 \begin{tabular}{cccccccccc}
   \toprule
    \multirow{2}{*}{Method} &   
    \multirow{2}{*}{Time(ms)} &
	\multirow{2}{*}{Size(kb)} &
    \multirow{2}{*}{Accuracy(\%)} & \\ \\
 \midrule
    \multirow{1}{*} {Two stage TL+SD}  & 150 & 1500 & 95.4\\
 \midrule
    \multirow{1}{*} {TeLCoS}  & 60 & 800 & 94.8\\         
\bottomrule
\end{tabular}
\end{table}

\begin{figure}[t!]
\includegraphics[width=80mm]{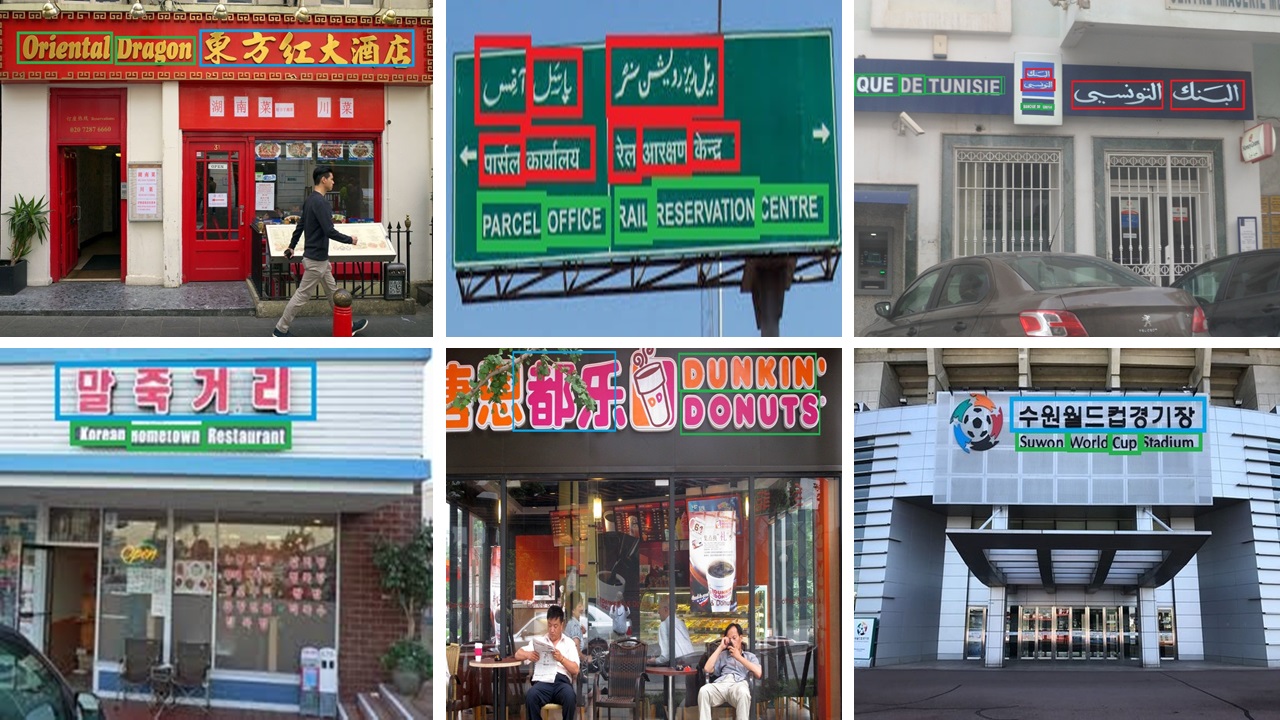}
\label{fig:label}
\centering
\caption{Sample Results. The script cluster is identified using Green, Blue, Red bounding box for the text instances corresponding to Latin, CJK, Others respectively.}
\centering
\end{figure}

\section{Conclusion}
In this work, we propose TeLCoS, an efficient end-to-end framework that combines Text Localization and High-level Script Clustering into a unified network. By sharing of convolution modules, we allow the network to adaptively learn the weights. The multi-task learning helps us to reduce the computational overhead of performing separate script identification for the text instances present in any image. Thus, achieving real-time speed for text localization and script identfication task. We introduce a novel method of channel pruning for compression of models, making them suitable for resource constrained devices. Comparison with state of the art results shows that our method performs well as compared to previous results in terms of performance and efficiency. 

\end{document}